\pdfoutput=1

\documentclass[11pt]{article}

\usepackage{acl}

\usepackage{times}
\usepackage{latexsym}

\usepackage{amssymb}
\usepackage{amsmath}
\usepackage{amsfonts}
\usepackage{booktabs}
\usepackage{graphicx}
\usepackage{multirow}
\usepackage{inconsolata}
\usepackage{algpseudocode}
\usepackage{algorithm}

\algrenewcommand\algorithmicrequire{\textbf{Input:}}
\algrenewcommand\algorithmicensure{\textbf{Output:}}
\usepackage[T1]{fontenc}

\usepackage[utf8]{inputenc}

\usepackage{microtype}

\usepackage{graphicx}

\newcommand{\qb}{\boldsymbol{q}}

\newcommand{\xb}{\boldsymbol{x}}
\newcommand{\rb}{\boldsymbol{r}}
\newcommand{\vb}{\boldsymbol{v}}

\definecolor{tabhighlight}{HTML}{e5e5e5}

\title{Graph Elicitation for Guiding Multi-Step Reasoning \\in Large Language Models}

\author{Jinyoung Park$^{1}$\thanks{Part of this work was done during an internship at Amazon AGI.}, Ameen Patel$^{2}$\thanks{Work was done while at Amazon AGI.}, Omar Zia Khan$^3$, Hyunwoo J. Kim$^{1}$\thanks{Co-corresponding authors.}, Joo-Kyung Kim$^{3\ddagger}$\\
$^1$Korea University, $^2$Together AI, $^3$Amazon AGI\\
{\tt\small \{lpmn678, hyunwoojkim\}@korea.ac.kr}\\
{\tt\small ameen@together.ai}\\
{\tt\small \{ozkhan, jookyk\}@amazon.com}\\
}

\begin{document}
\maketitle
\begin{abstract}

Chain-of-Thought (CoT) prompting along with sub-question generation and answering has enhanced multi-step reasoning capabilities of Large Language Models (LLMs).
However, prompting the LLMs to directly generate sub-questions is suboptimal since they sometimes generate redundant or irrelevant questions.
To deal with them, we propose a GE-Reasoning method, which directs LLMs to generate proper sub-questions and corresponding answers.
Concretely, given an input question, we first prompt the LLM to generate knowledge triplets, forming a graph representation of the question.
Unlike conventional knowledge triplets, our approach allows variables as head or tail entities, effectively representing a question as knowledge triplets. 
Second, for each triplet, the LLM generates a corresponding sub-question and answer along with using knowledge retrieval. If the prediction confidence exceeds a threshold, the sub-question and prediction are incorporated into the prompt for subsequent processing.
This approach encourages that sub-questions are grounded in the extracted knowledge triplets, reducing redundancy and irrelevance.
Our experiments demonstrate that our approach outperforms previous CoT prompting methods and their variants on multi-hop question answering benchmark datasets.

\end{abstract}

\section{Introduction}
Large Language Models~(LLMs)~\cite{brown2020language,ouyang2022training,chowdhery2022palm,zhang2022opt,touvron2023llama,touvron2023llama2,meta2024introducing} have shown remarkable performance on various natural language processing tasks even without fine-tuning for the target tasks.
Specifically, Chain-of-Thought (CoT) prompting approach~\cite{wei2022chain,kojima2022large} has improved the reasoning capability of the LLMs by 
generating intermediate rationales before making the final answer.
Although CoT prompting and the variants have shown better performance on various reasoning tasks~\cite{wei2022chain,wang2022self,jung2022maieutic,liu2022rainier}, they have difficulty in answering complex multi-hop questions~\cite{press2023measuring} with two problems: lack of knowledge and properly decomposing the question.



To deal with them, one approach is generating relevant sub-questions that are easier to answer than the original question and answering them. Using them as the context information has been shown effective to improve the reasoning performance \cite{patel2022question,radhakrishnan2023question,lyu2023faithful,kim2023kggpt,ling2023deductive,qi2023art}.
However, previous methods sometimes generate irrelevant, redundant, or insufficient sub-questions since they mostly rely on in-context learning (ICL) \cite{brown2020language} with raw text exemplars to decompose the original question to the sub-questions without concrete guidances.

To address the issues of the sub-question generation, we propose a graph-guide prompting method, which leads LLMs to generate proper sub-questions and the answers by using graphs elicited from the question and the contexts.
Our method first construct a question graph by leveraging LLM prompting with in-context learning to extract knowledge triplets from the question. Differently from conventional knowledge triplets, our approach let each triplet include variables as the head or the tail entities. This relaxed triplet representation facilitates represent question sentences as triplets.
Second, for each triplet, the LLM generates a corresponding sub-question grounded to the triplet. Then, the LLM answers it along with intermediate rationales and external knowledge retrieval.
Those sub-question and the answer pairs with low answer confidences are filtered out by confidence thresholding.
We repeat the intermediate sub-question/rationale/answer generation step, and the filtering step until reaching the final answer to the original question or the maximum number of reasoning steps.

Using knowledge triplets in the sub-question generation has three benefits.
First, since each sub-question is grounded by a knowledge triplet extracted from the question, the sub-questions are highly likely to be relevant to the original question.
Second, as each sub-question is generated from a distinct triplet, the sub-questions are not redundant.
Third, we can better track how each sub-question is grounded and composed.
Additionally, we allow representing triplet entities as variables, which acts as the entity placeholders and facilitates representing question sentences as the triplets. 
Allowing variables in the knowledge triplets resembles first-order logic (FOL) \cite{barker2011lan} representations, which are useful for specific tasks such as claim verification \cite{wang2023explainable}, but we do not require strict formal representations or external theorem provers for the reasoning process \cite{wang2023explainable,olausson2023linc,pan2023logiclm}.

For each generated sub-question, the LLM generates the answer. We filter out uninformative sub-question/answer pairs if the answer confidence is below a threshold, similarly to \cite{jiang2023active}.
In the open-book settings, where we can retrieve external knowledge, we allow retrieving relevant paragraphs by using the sub-question as the query so that the answer can be better generated leveraging the retrieved information.

Some previous works also leverage extracted entities or triplets. \citet{sun2023thinkongraph,jiang2023structgpt} uses extracted entities for traversing given external knowledge graphs in knowledge-base question answering.
\citet{fu2021decomposing} uses entities and relations for sub-question generation, and it is evaluated on 2WikiMultiHopQA~\cite{ho2020constructing}, but their approach is less flexible since it requires fine-tuning of multiple different components, the question type should be separately estimated, and the final answer is limited as an aggregation of the sub-questions' answers.
Recent works~\cite{li2023leveraging,liu2024era} extract entities and their relations from the knowledge documents and augment them to the input prompt for reasoning.
Different from them, we elicit a graph with variables from the input question and leverage it to decompose the complex question into multiple simple and relevant sub-questions.

We evaluate the effectiveness of our proposed methods on three multi-hop reasoning benchmark datasets: 2WikiMultihopQA~\cite{ho2020constructing}, MuSiQue~\cite{trivedi2022musique}, and Bamboogle~\cite{press2023measuring}.
Our experiments are conducted with 
Llama-3~\cite{meta2024introducing}, which is a widely used open source LLM, with two model sizes (8B and 70B), and GPT-3.5 Turbo \cite{ouyang2022training}, which is a popular proprietary LLM.
From the experiments, our method shows the best performance compared to the other prompting methods on top of the LLMs in both the closed-book (no retrieval) and the open-book (knowledge retrieval) settings.

Our main contributions are as follows:
\begin{itemize}
    \item We propose a GE-Reasoning method that elicits knowledge triplets of the questions and utilize them for generating relevant and distinctive sub-questions.
    \item We propose an in-context learning method for extracting knowledge triplets with variable entities, which allows suitable triplet representations of the input questions.
    \item We present retrieval augmented generation with structural knowledge refinement that filters out irrelevant knowledge information.
    \item Our extensive experiments demonstrate that our proposed approach outperform the baselines on three multi-hop question answering benchmark datasets: 2WikiMultihopQA, MuSiQue, and Bamboogle. 
\end{itemize}
\vspace{-3mm}
\section{Related Works}

\subsection{Prompts for Multi-Step Reasoning}
Chain-of-thoughts prompting~\cite{wei2022chain,chowdhery2022palm,kojima2022large} has been successfully applied to various reasoning tasks by providing the reasoning steps in the demonstrations.
The other approach for multi-step reasoning~\cite{jung2022maieutic,press2023measuring,gao2023pal,creswell2023selection,parisi2022talm,schick2023toolformer} is applying symbolic functions to the prompting.
Further, least-to-most prompting~\cite{zhou2023least} proposes a multi-stage prompting approach where one prompt is designed for generating sub-questions, and the other prompt is used for answering the sub-questions.
In addition, some works~\cite{hu2023code,pan2023factchecking,lyu2023faithful} explore code-based approaches with external compilers to execute the code. 
Different from these works, our prompting method provides explicit guidance with elicited graphs to the LLMs to reason for the complex multi-hop questions.
Additionally, we iteratively generate the intermediate rationales and then verify them to reach the correct answer from the question without external programs.

\subsection{Prompting with Knowledge Retrieval}
Prompting has also been widely applied to open-book question answering tasks requiring external knowledge information. ~\citet{lazaridou2022internet,sun2023recitation,yu2022generate} use prompting methods for single retrieval for each question, which is suboptimal for knowledge intensive multi-hop questions.
Self-Ask~\cite{press2023measuring} is proposed to improve the reasoning capability of LLMs by decomposing the question into sub-questions and simply answers the sub-questions using Google Search API.
However, this approach is not based on multi-step reasoning, which is still with limited capabilities.
ReAct~\cite{yao2023react} prompting generates a sequence of reasoning steps and action steps, but its performance highly depends on the scale of language models and it requires fine-tuning to outperform conventional chain-of-thought prompting methods on multi-hop question answering in the open domain setting.
IRCoT~\cite{trivedi2023interleaving} uses knowledge retrieval given the intermediate thought as a query.
Therefore, it sometimes replaces an accurate rationale with an incorrect rationale influenced by the noisy knowledge.
Compared to these related works, our method shows the effectiveness for diverse sizes of LLMs on multi-hop question answering tasks in the open-domain settings.
\section{Graph Elicitation for Guiding Multi-Step Reasoning}
\begin{figure*}[t] 
\centering
\includegraphics[width=1.0\textwidth]{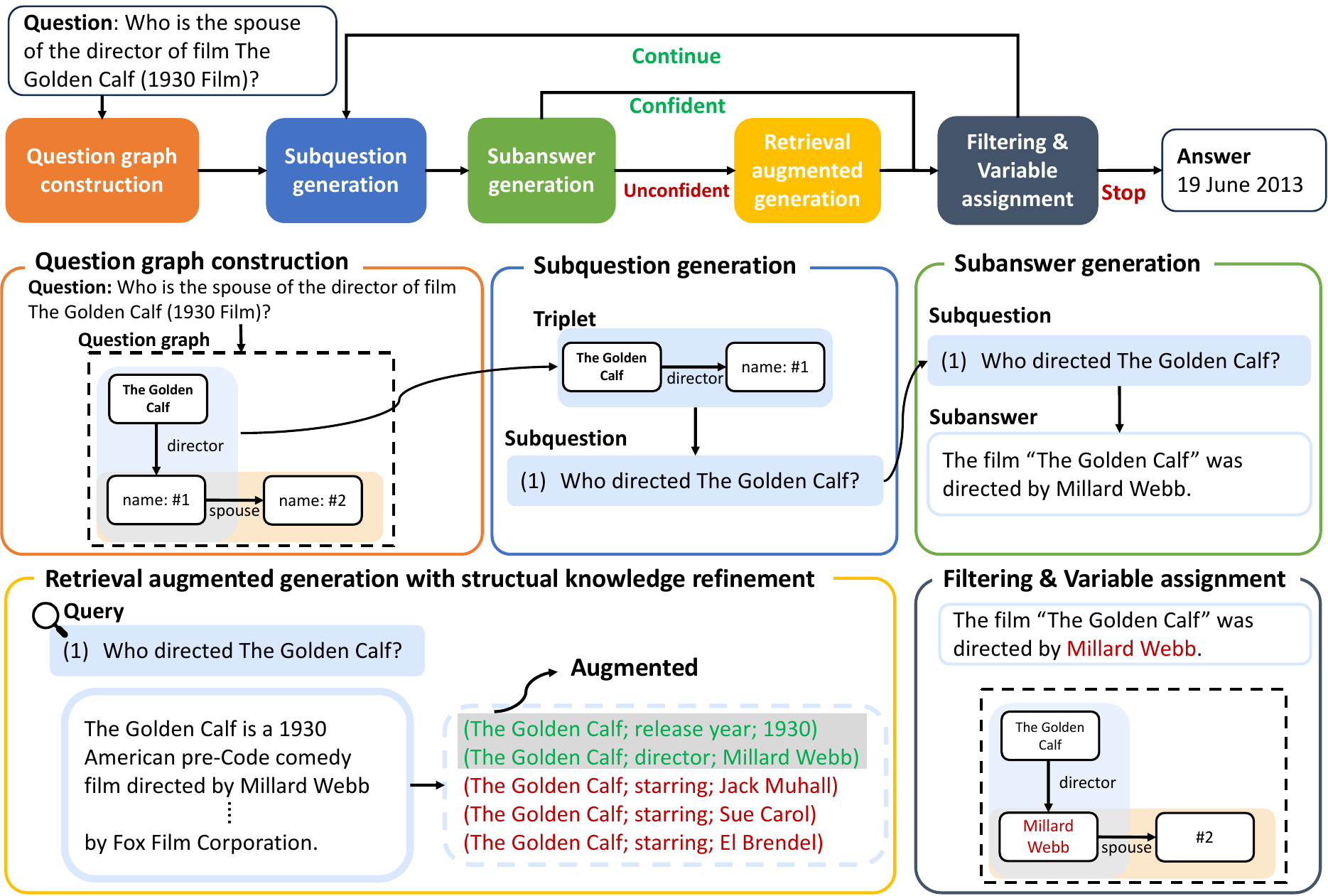}
\caption{
The overview of our graph-guided prompting method for the complex reasoning task. We first construct the question graph by extracting the triplets from the question with in-context learning. Then, we sequentially repeat the following steps: (1) sub-question generation step~(Sec.~\ref{sec:subq}) that generates an intermediate sub-question based on a triplet with one variable~(\texttt{(The Golden Calf; director; name: \#1)}) of the question graph. (2) sub-answer generation step~(Sec.~\ref{sec:suba}) that generates the sub-answer by answering the previously generated sub-question. 
(3) retrieval augmented generation step~(Sec.~\ref{sec:ret}) that retrieves the knowledge given the sub-question as a query and then extracts triplets followed by filtering them to augment informative ones, based on the confidence thresholding. 
(4) variable assignment step~(Sec.~\ref{sec:verify}) that fills the variable based on the sub-answer.
If there are no remaining question triplets with single variables and the repetition limit is reached, we stop the iteration.
After the end of the iterations, the final answer to the original question is generated.
}
\label{fig:main_fig}
\vspace{-12pt}
\end{figure*}


\begin{figure*}[t] 
\centering
\includegraphics[width=1.0\textwidth]{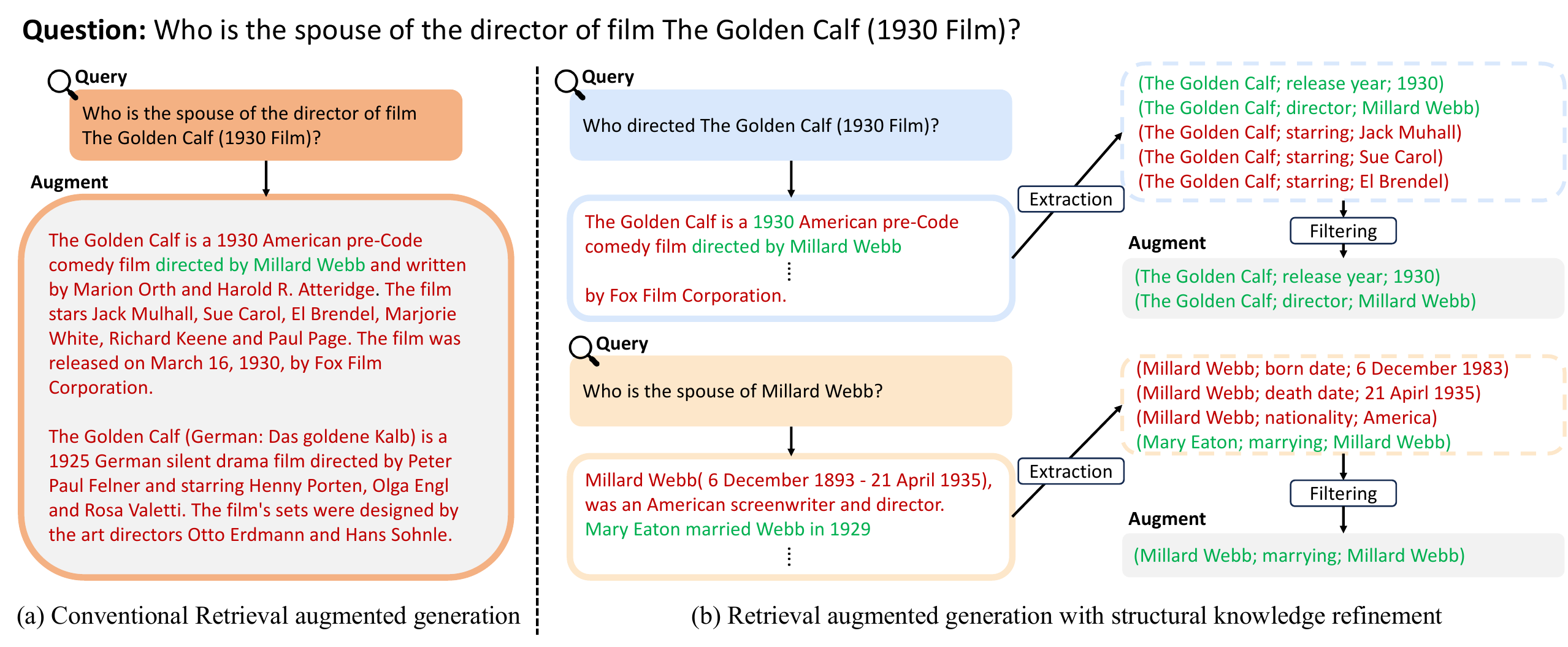}
\caption{
A comparison of conventional and our retrieval augmented generation approaches. 
In the conventional approaches, the input query is compound and complex to limit retrieving the full knowledge sufficient to reason over multi-hop question.
It also augments all the information, including irrelevant ones, to the input prompt without filtering them.
Our retrieval augmented generation approach uses a sub-question of each step as an input query for the retrieval system.
After the retrieval, knowledge triplets are extracted from the knowledge paragraphs, and then relevant triplets are augmented to the input prompt after the filtering. 
}
\label{fig:ret_fig}
\vspace{-12pt}
\end{figure*}


The goal of our \textbf{G}raph \textbf{E}licitation for guiding multi-step \textbf{Reasoning}~(\textbf{GE-Reasoning}) is to solve complex multi-hop question-answering by decomposing the complex question into multiple sub-questions and generating the sub-answers with guidance based on elicited graphs.
The overview of our GE-Reasoning is depicted in Figure~\ref{fig:main_fig}. 
Concretely, we start with constructing a question graph by eliciting a graph from the question~(\textbf{Step1. Question graph construction~(Sec.~\ref{sec:question_construction})}).
Then, we iterate the following steps.
First, we generate a sub-question $\qb^{(j)}$ based on one of the extracted triplets to obtain the information required to answer the input question $\xb$ by referring to the question graph (\textbf{Step2. Sub-question generation~(Sec.~\ref{sec:subq})}).
After generating the sub-question, we predict its answer $\rb^{(j)}$ by prompting LLMs~(\textbf{Step3. Sub-answer generation~(Sec.~\ref{sec:suba})}) along with \textbf{Step4. retrieval augmented generation~(Sec.~\ref{sec:ret})} if needed.
Next, we evaluate the confidence of the subanswer and use it to fill the variable entity~(\textbf{Step5. Filtering and Variable assignment~(Sec.~\ref{sec:verify})}).
If there exist no remaining question triplets with variables or the maximum number of the repetitions is reached, we stop the iterative generation and predicts the final answer of the input question.
The pseudocode of GE-Reasoning is in Algorithm~\ref{alg}.



\subsection{Question Graph Construction.}
\label{sec:question_construction}
We construct the question graph $\mathcal{G}_q = \left(\mathcal{V}_q, \mathcal{R}_q, \mathcal{T}_q \right)$ by extracting a set of triplets from the question $q$, where $\mathcal{V}_q$ and $\mathcal{R}_q$ are sets of nodes and relations, respectively. 
$\mathcal{T}_q$ denotes a set of triplets, $\boldsymbol{t} = \left(\vb_h, \rb, \vb_t\right)$, where $\vb_h,\vb_t$ are head and tail nodes, respectively, and $\rb$ is the relation between the two nodes. 
This question graph is a \emph{heterogeneous} graph that consists of various types (relations) of edges connecting entities, which can be represented as (head, relation, tail) triplets.
One way to construct the question graph is using relation extraction models~\cite{fu2019graphrel,fu2021decomposing,melnyk2022knowledge}. 
However, they require additional training step and their graph constructions do not generalize well beyond question sentences and diverse datasets.

To address these limitations, we introduce an in-context learning method to harness the power of LLMs.
Given a question sentence $\boldsymbol{x}$, we prompt the language model to extract triplets from the sentence as follows:
\begin{equation}
\label{eq:graph_construction}
\mathcal{G}_{\xb} = \text{LM}\left(\mathcal{E}_G, \boldsymbol{x} \right),
\end{equation}
where $\mathcal{G}_{\xb}$ the question graph represented by a set of triplets and $\mathcal{E}_G = \left\{(\boldsymbol{x}_i, \mathcal{G}_i) \right\}_{i=1}^{|\mathcal{E}_G|}$ is a set of exemplars consisting of pairs of input $\boldsymbol{x}_i$ and the corresponding graph $\mathcal{G}_i$.
The input prompt for the question graph construction is in Table~\ref{tab:pr2}.
For example, given the question ``Who is the spouse of the director of film The Golden Calf (1930 Film)", we extract the token sequence of triplets $\mathcal{G}_{\xb}=\{$  \texttt{(The Golden Calf; director; \texttt{name: \#1}), (\texttt{name: \#1}; spouse; \texttt{name: \#2})} $\}$ by Eq.~\eqref{eq:graph_construction}, where \texttt{name: \#1, name: \#2} are variables with their type denoting the entities required to be answered via sub-questions.
The types provide hints about the variables, helping the model to comprehend them better.
Once all the knowledge triplets are generated, we filter out invalid triplets such as those without any variable entities or with invalid formats.

\subsection{Sub-Question Generation}
\label{sec:subq}
We generate sub-questions $\boldsymbol{q}^{(j)}$ with concrete guidance by the generated graph ${\mathcal{G}}_{\boldsymbol{x}}$. 
While the previous sub-question generation works such as \citet{radhakrishnan2023question,yoran2023answering} generate subquestions only dependent on in-context demonstrations, we explicitly guide the model to generate subquestions with the graph structure $\mathcal{G}_{\boldsymbol{x}}$ (\textit{i.e.}, knowledge triplets).
Specifically, we first sample a triplet with one variable from the graph $\mathcal{G}_{\boldsymbol{x}}$ and construct a candidate set $\mathcal{C}_{\boldsymbol{x}} \subseteq \mathcal{G}_{\boldsymbol{x}}$. 
Then, the LLM generates a sub-question $\boldsymbol{q}^{(j)}$ based on the question triplet $\boldsymbol{t}^{(j)} \in \mathcal{C}_{\boldsymbol{x}}$, which can be formulated as:
\begin{equation}
    \boldsymbol{q}^{(j)} = \text{LM}\left(\mathcal{E}_{t \rightarrow q}, \boldsymbol{t^{(j)}} \right),
\end{equation}
where $\mathcal{E}_{t \rightarrow q}=\left\{\left(\boldsymbol{t}_i, \boldsymbol{q}_i \right) \right\}_{i=1}^{\lvert \mathcal{E}_{t \rightarrow q}\rvert}$ is a set of exemplars consisting of pairs of triplet $\boldsymbol{t}_i$ and the corresponding sub-question $\boldsymbol{q}_i$.
The triplets with two variables are not used for sub-question generation, but they can be sub-question generatable if one of the variables are assigned later as described in Section~\ref{sec:verify}.
The graph-based guidance facilitates generating relevant and distinct sub-questions.
Our experimental result in Section~\ref{sec:additional} shows that the proposed approach effectively suppresses generating irrelevant sub-questions.

\subsection{Sub-Answer Generation}
\label{sec:suba}
Given each sub-question $\boldsymbol{q}^{(j)}$ 
along with previously generated sub-questions and sub-answers, the LLM generates the corresponding answer $\boldsymbol{r}^{(j)}$ :
\begin{equation}
    \boldsymbol{r}^{(j)} = \text{LM}\left(\mathcal{E}, \boldsymbol{x}, \boldsymbol{q}^{(1)}, \boldsymbol{r}^{(1)}, \dots \boldsymbol{q}^{(j)} \right).
\end{equation}
We use stochastic decoding with temperature 0.6 for the diverse answers during the iteration process.
Since each sub-question $\boldsymbol{q}^{(i)}$ is simpler and shorter than the original question, it is easier for the LLM to predict the correct answer. 

However, the LLM may still provide uncertain sub-answers due to the lack of knowledge. 
Motivated by \citet{jiang2023active}, we maintain the generated sub-answer if the confidence is above a threshold.
Otherwise, we use retrieval augmented generation, where knowledge retrieval is leveraged for the sub-answer generation, as described in the next section.

\subsection{Retrieval Augmented Generation with Structural Knowledge Refinement}
\label{sec:ret}
A common approach for the retrieval augmented generation is single-level retrieval, which directly uses the input question $\boldsymbol{x}$ as the query for the retrieval system and then uses both the input and the retrieved knowledge documents $\mathcal{D}_{\boldsymbol{x}}=\text{Ret}\left(\boldsymbol{x}\right)$ to generate the answer~\cite{chen2017reading,guu2020retrieval,lewis2020retrieval}.
However, if the question is compound or requires multi-step reasoning, where important questions emerge amidst the reasoning process, a single retrieval might return irrelevant content or miss crucial information.

To deal with these issues, our retrieval method uses multi-level retrieval with sub-questions and refines the retrieved knowledge.
Existing multi-level retrieval approaches such as IRCoT~\cite{trivedi2023interleaving} use a sentence of the previous reasoning step as a query for the retrieval system.
However, the previous sentence may be unrelated to the current reasoning step, which results in the retrieving irrelevant content.
Thus, we use the generated sub-question $\boldsymbol{q}^{(j)}$ (\textit{e.g.}, `Who directed The Golden Calf (1930 Film)?') as the query instead of the complex input question or the sentence of the previous reasoning step.
The sub-question reflects the information needed in the current reasoning step, facilitating the retrieval of informative knowledge. 

Since we retrieve paragraphs, those retrieved contents may contain both relevant and irrelevant information to the sub-question.
Therefore, we extract only relevant information from the paragraphs to help answering the sub-questions.
The overall procedure is depicted in Figure~\ref{fig:ret_fig}.
Specifically, we extract all possible knowledge triplets from each retrieved raw paragraph using the prompt in Table~\ref{tab:pr4}.
Then, we filter out irrelevant triplets using the prompt in Table~\ref{tab:pr5}.
Finally, as the outcome of the retrieval, we append the remaining triplets to the input prompt instead of the raw knowledge paragraphs.
Some recent works~\cite{li2023leveraging,liu2024era} also extract knowledge triplets from the additional context, but they do not consider the relevance between the extracted triplets and the input question, and use all the triplets including noisy ones.
Different from them, we use only informative triplets as additional contexts to the reasoning.

\subsection{Filtering and Variable Assignment} 
\label{sec:verify}
The filtering step evaluates the confidence of the sub-answer.
If the LLM generates low-confident answers even when retrieval augmentation is used, we filter out the current sub-question and the corresponding sub-answer.
Note that even if the current sub-question and the answer are filtered out due to low confidence, the stochastic sub-answer generation may produce high confident answers in subsequent iterations.

If the answer has a high-confidence, we assign the answer to the variable of the triplet used for generating the sub-question.
If the same variable exists in other triplets, we also update them accordingly.
For example, in Figure~\ref{fig:main_fig}, `Millard Webb` is assigned to the variable entity $\texttt{name: \#1}$ after getting the answer of the sub-question `Who directed The Golden Calf?'.
Since $\texttt{name: \#1}$ also exists in $\texttt{(name: \#1; spouse; name: \#2)}$, the triplet is changed to $\texttt{(Millard Webb; spouse; name: \#2)}$, which becomes eligible for sub-question generation.

\subsection{Iterative Generation}
We stop generating sub-questions and their answers when no remaining question triplets with single variables exist, or the repetition limit\footnote{10 in our experiments.} is reached.
Once the iteration is over, we generate the answer for the original question with the following instruction at the end of the prompt: ``So the answer is \texttt{answer}''.

{\renewcommand{\arraystretch}{0.92}%
\begin{table*}[!ht]
    \setlength{\tabcolsep}{3pt} 
    
    \centering
    \resizebox{\textwidth}{!}{
    \begin{tabular}{lccccccccccccccc}
        \toprule
            & \multicolumn{5}{c}{Llama3-8B} & \multicolumn{5}{c}{Llama3-70B} & \multicolumn{5}{c}{GPT-3.5 Turbo} \\
          \cmidrule(l{3pt}r{3pt}){2-6}
          \cmidrule(l{3pt}r{3pt}){7-11}
          \cmidrule(l{3pt}r{3pt}){12-16}
          & \multicolumn{2}{c}{2WikiMHQA}&\multicolumn{2}{c}{MusiQue} & \multicolumn{1}{c}{Bamboo}& \multicolumn{2}{c}{2WikiMHQA}&\multicolumn{2}{c}{MusiQue} & \multicolumn{1}{c}{Bamboo} & \multicolumn{2}{c}{2WikiMHQA}&\multicolumn{2}{c}{MusiQue} & \multicolumn{1}{c}{Bamboo}\\
          \cmidrule(l{3pt}r{3pt}){2-3}
          \cmidrule(l{3pt}r{3pt}){4-5}
          \cmidrule(l{3pt}r{3pt}){6-6}
          \cmidrule(l{3pt}r{3pt}){7-8}
          \cmidrule(l{3pt}r{3pt}){9-10}
          \cmidrule(l{3pt}r{3pt}){11-11}
          \cmidrule(l{3pt}r{3pt}){12-13}
          \cmidrule(l{3pt}r{3pt}){14-15}
          \cmidrule(l{3pt}r{3pt}){16-16}
          Methods    & EM & F1     & EM & F1 & EM  & EM & F1     & EM & F1 & EM  & EM & F1     & EM & F1 & EM  \\
         \midrule
         No retrieval& 31.6 & 38.94 & 11.4 & 21.11 & 45.4 &48.0 & 56.41 & 21.4 & 32.18 & 64.8&37.0 & 45.28 & 15.2 & 25.49 & 56.6\\
         One retrieval & 35.0 & 44.66  & 16.2 & 24.99 & 52.8 & 53.8 & 65.75 &  25.2 & 35.92 & 67.2 & 40.6 & 50.80 &  16.4 & 25.75 & 57.0\\
         Verify-and-Edit &  36.4 & 43.00   & 18.8 & 27.54 & 52.6 &  55.8 & 66.17  & 30.0 & 41.41 & 66.8 &  42.8 & 52.54  & 16.6 & 25.91 & 58.0\\
         FLARE & 41.8 & 50.70  & 22.6 & 31.72 & 54.0 & 65.2 & 72.62  & 32.0 & 42.86 & 69.0 & 50.8 & 61.31  & 19.4 & 31.33 & 57.2\\
         IRCoT &  45.6 & 57.35  & 22.6 & 31.94 & 53.2 &  63.8 & 73.39 &  29.6 & 40.96 & 69.6 &  48.2 & 58.53 &  17.8 & 29.06 & 57.6 \\ 
         ERA-CoT &  35.2 & 42.93  & 18.0 & 26.64 & 52.8 &  54.8 & 65.78 &  26.2 & 37.60 & 66.2 &  42.0 & 51.67 &  16.4 & 26.02 & 56.2\\ 
         \midrule
         \textbf{GE-Reasoning~(\textit{Ours})} & \textbf{53.0} & \textbf{60.99}  & \textbf{23.8} & \textbf{33.06} & \textbf{55.8} & \textbf{66.2} & \textbf{77.54} &  \textbf{33.0} & \textbf{43.95} & \textbf{72.0} & \textbf{52.8} & \textbf{64.03} &  \textbf{23.2} & \textbf{32.83} & \textbf{59.0}\\
        \bottomrule
    \end{tabular}
    }
    \vspace{-2mm}
    \caption{Performance comparisons on multi-hop question answering datasets in the open-book setting. 
    }
    \label{tab:ret}
    \vspace{-6mm}
\end{table*}
}
\section{Experiments}
\subsection{Experimental Setup}
We evaluate the effectiveness of our proposed methods on \textcolor{black}{three} multi-hop question answering benchmark datasets: \textbf{2WikiMultihopQA} (2WikiMHQA)~\cite{ho2020constructing}, \textbf{MuSiQue}~\cite{trivedi2022musique}, and \textbf{Bamboogle} (Bamboo)~\cite{press2023measuring}.
For the open-domain setting, we use the set of paragraphs provided in 2WikiMHQA and MuSiQue to curate an external knowledge corpus following the other existing works~\cite{trivedi2023interleaving}.
For Bamboogle dataset, we use a retrieval based on Google search following existing works~\cite{yoran2023answering,zhao-etal-2023-verify}.
In addition, we follow ~\citet{trivedi2023interleaving,jiang2023active} to randomly subsample 500 questions out of the dev set for 2WikiMHQA and MuSiQue datasets.
We use all 125 questions for Bamboo dataset.
We provide 6 exemplars for the in-context learning to predict the answer on all the datasets.
We evaluate the performance of the approaches with the answer-level exact match~(EM) and token-level F1.
\subsection{Models}

We experiment with open source Llama-3 (8B and 70B)~\cite{meta2024introducing} and proprietary GPT 3.5 Turbo \cite{ouyang2022training} as the base LLMs.
For the knowledge retrieval, we employ BM25~\cite{robertson2009probabilistic} implemented with Elasticsearch and use the top-2 retrieved documents following other RAG works~\cite{jiang2023active}.

\subsection{Baselines}
In the open-book setting, which leverages the external knowledge, we use the following baselines:
\textbf{No retrieval}: predicting the answer using CoT prompting without any external knowledge.
\textbf{One retrieval}: predicting the answer using CoT prompting with the context retrieved with the input question as the query.
\textbf{Verify-and-Edit}~\cite{zhao-etal-2023-verify}: generating the reasoning steps and then predicting the answer after editing inaccurate sentences with the retrieved knowledge.
\textbf{FLARE}~\cite{jiang2023active}: actively retrieving the knowledge context based on confidence and predicting an answer with the retrieved context.
\textbf{IRCoT}~\cite{trivedi2023interleaving}: interleaving the retrieval with sentences and answer generation with the retrieved context. 
\textbf{ERA-CoT}~\cite{liu2024era}: capturing relationships between entities and adding the relationships to the input prompt for better reasoning.
While \citet{liu2024era} uses ERA-CoT with gold knowledge as the additional contexts, we evaluate ERA-CoT with retrieved knowledge for fair comparisons with the other approaches.

To validate the effectiveness of our proposed prompting method without knowledge retrieval (\textit{i.e.,} closed book setting), we use the following baselines:
    \textbf{Chain-of-Thoughts}~\cite{wei2022chain}: predicting the answer with the reasoning steps with ICL exemplars.
    \textbf{Zero-Plus-Few-Shot CoT}~\cite{kojima2022large}: including ``Let's think step by step.'' before the reasoning steps of CoT.
    \textbf{Self-Consistency}~\cite{wang2022self}: sampling five reasoning paths with a decoding temperature of 0.7 and using majority voting to get the answer.
    \textbf{Self-ask}~\cite{press2023measuring}: sequentially asking questions until reaching the final answer.  

\subsection{Experimental Results}
    
          

\begin{table}[!t]
    
    \centering
    \resizebox{\columnwidth}{!}{
    \begin{tabular}{llcccc}
        \toprule
           & & \multicolumn{2}{c}{2WikiMHQA}&\multicolumn{2}{c}{MusiQue} \\
           \cmidrule(l{3pt}r{3pt}){3-4} \cmidrule(l{3pt}r{3pt}){5-6}
            Ret. Query &  Subq Gen. & EM & F1 &EM & F1 \\
         \midrule 
          Input question & w/o Subq Gen.  & 35.0 & 44.66 & 16.2 & 24.99 \\
          Input question & w/o G-Guidance &35.8 & 44.22& 16.4&24.70 \\
          Input question & with G-Guidance  & \textbf{37.6}& \textbf{47.61} & \textbf{17.0} & \textbf{26.13}\\
          \midrule
          Subquestion & w/o G-Guidance  & 49.2 & 57.32 & 22.4 & 31.76\\ 
          Subquestion& with G-Guidance & \textbf{53.0}& \textbf{60.99}& \textbf{23.8} & \textbf{33.06} \\
        \bottomrule
    \end{tabular}
    }
    \vspace{-2mm}
    \caption{Performance comparison based on the retrieval query types and subquestion generation methods on multi-hop question answering datasets with Llama3-8B. (Ret. Query: Retrieval query type, Subq. Gen.: Subquestion generation).
    }
    \label{tab:ablation}
    \vspace{-2mm}
\end{table}
We evaluate our proposed methods using Llama3-8B, Llama3-70B, and GPT-3.5 Turbo in Table~\ref{tab:ret}. 
From the table, our GE-Reasoning shows the best performance compared to the other baseline prompting methods on all the datasets with various LLMs.
This results indicate that our prompting methods are widely applicable to diverse LLMs with different sizes for multi-hop QA tasks.

\subsection{Additional Experimental Results.}
\label{sec:additional}

\begin{table}[!t]
    
    \centering
    \resizebox{\columnwidth}{!}{
    \begin{tabular}{cccccc}
        \toprule
           & & \multicolumn{2}{c}{2WikiMHQA}&\multicolumn{2}{c}{MusiQue} \\
           \cmidrule(l{3pt}r{3pt}){3-4} \cmidrule(l{3pt}r{3pt}){5-6}
            Knowl. Refine. &  Filter. & EM & F1 &EM & F1 \\
         \midrule 
           &   & 52.0 &58.68 & 22.8& 31.63\\
           \checkmark &   &\textbf{53.4} &60.41 & 23.2& 32.18\\
          \checkmark & \checkmark  &53.0 &\textbf{60.99} &\textbf{23.8} &\textbf{33.06} \\
          
        \bottomrule
    \end{tabular}
    }
    \vspace{-2mm}
    \caption{Ablation studies of our GE-Reasoning on multi-hop question answering datasets with Llama3-8B. (Knowl. Refine.: Knowledge Refinement, Filter.: Filtering).
    }
    \label{tab:abl2}
    \vspace{-6mm}
\end{table}
\begin{table}[!ht]
    
    \centering
    \resizebox{\columnwidth}{!}{
    \begin{tabular}{lcccc}
        \toprule
           & \multicolumn{2}{c}{2WikiMHQA}&\multicolumn{2}{c}{MusiQue} \\
           \cmidrule(l{3pt}r{3pt}){2-3} \cmidrule(l{3pt}r{3pt}){4-5}
          Methods    & EM & F1     & EM & F1  \\
         \midrule 
         Chain-of-Thoughts      &31.6 & 38.94 & 11.4 & 21.11 \\
         Zero-Plus-Few-Shot CoT &34.2 & 40.09 & 11.4 & 20.57 \\
         Self-Consistency       & 33.8 & 40.87 & 11.6 & 21.21 \\
         Self-ask               &33.4 & 39.47 & 13.0 & 21.64 \\
         \midrule
         \textbf{GE-Reasoning~(\textit{ours})} & \textbf{36.4} & \textbf{42.75}& \textbf{15.4} & \textbf{24.91} \\
        \bottomrule
    \end{tabular}
    }
    \vspace{-2mm}
    \caption{Performance comparison on multi-hop question answering datasets using Llama-3-8B without knowledge retrieval.
    }
    \vspace{-6mm}
    \label{tab:noret}
\end{table}

\begin{table}[!ht]
    \small
    
    \centering
    \begin{tabular}{lccc}
        \toprule
           & \multicolumn{1}{c}{Graph quality}&\multicolumn{2}{c}{QA} \\
           \cmidrule(l{3pt}r{3pt}){2-2} \cmidrule(l{3pt}r{3pt}){3-4}
          Methods    & Acc.     & EM & F1  \\
         \midrule 
         w/o type   &95.4 & 52.2 & 60.16 \\
         with type & \textbf{96.0} & \textbf{53.0} & \textbf{60.99} \\
        \bottomrule
    \end{tabular}
    \vspace{-2mm}
    \caption{Evaluation of graph quality and question answering with and without using entity type when constructing a question graph on 2WikiMHQA dataset using Llama-3 8B.
    }
    \label{tab:graphqual}
    \vspace{-8mm}
\end{table}

\begin{table*}[!t]
    \scriptsize
    

    \centering
    \begin{tabular}{p{2.8cm}|p{6.2cm}|p{5.4cm}}
        \toprule
        \textbf{Question \& Answer} &  \textbf{w/o G-guidance} & \textbf{with G-guidance~(Ours)} \\
        \midrule
          \textbf{Question: }Which film whose director is younger, Running For Grace or Irulum Oliyum?
          \newline \textbf{Answer: }Running For Grace
            &
         (1) Who directed Irulum Oliyum? \newline
Irulum Oliyum was directed by Puttanna Kanagal. \newline
(2) Who directed Running For Grace? \newline
Running For Grace was directed by David L. Cunningham. \newline
\textcolor{red}{(3) Who directed Who is Running?} \newline
\textcolor{red}{Who is Running was directed by Oxide Pang Chun.}\newline
(4) Who is the youngest director among \textcolor{red}{Oxide Pang Chun}, Puttanna Kanagal, and David L. Cunningham? \newline
\textcolor{red}{Oxide Pang Chun is the youngest director.} \newline
\textcolor{red}{So the answer is \textbf{Who is Running}.}
            &
(1) Who directed Irulum Oliyum? \newline
\textcolor{gray}{\tiny(Irulum Oliyum; director; name: \#1)}\newline
Irulum Oliyum was directed by Puttanna Kanagal. \newline
(2) Who directed Running For Grace? \newline
\textcolor{gray}{\tiny(Running For Grace; director; name: \#2)}\newline
Running For Grace was directed by David L. Cunningham. \newline
(3) Who is younger, Puttanna Kanagal or David L. Cunningham? \newline
David L. Cunningham is younger. \newline
So the answer is \textbf{Running For Grace.}\\
    \midrule
          \textbf{Question: }Who is Ahaziah Of Israel's maternal grandfather?
          \newline \textbf{Answer: }Ithobaal I
            &
         (1) Who is Ahaziah Of Israel's mother? \newline
Ahaziah Of Israel's mother is Jezebel. \newline
\textcolor{red}{(2) Who is Jezebel's husband?} \newline
\textcolor{red}{Jezebel's husband is Ahab, King of Israel.} \newline
(3) Who is \textcolor{red}{Ahab}'s father? \newline
\textcolor{red}{Ahab's father is Omri, King of Israel.} \newline
(4) Who is Ahaziah Of Israel's maternal grandfather? \newline
\textcolor{red}{Ahaziah Of Israel's maternal grandfather is Omri, King of Israel.} \newline
\textcolor{red}{So the answer is \textbf{Omri}}.
            &
(1) Who is Ahaziah Of Israel's mother? \newline
Ahaziah Of Israel's mother is Jezebel. \newline
(2) Who is Jezebel's father? \newline
Jezebel's father is Ithobaal I of Sidon. \newline
(3) Who is Ahaziah Of Israel's maternal grandfather? \newline
Ahaziah Of Israel's maternal grandfather is Ithobaal I of Sidon. \newline
So the answer is \textbf{Ithobaal I of Sidon}.
\\

        \bottomrule
    \end{tabular}
    \vspace{-2mm}
    \caption{Comparison on reasoning steps and answers generated by prompting without and with the graph guidance.
    }
    \label{tab:qual}
    \vspace{-5mm}
\end{table*}

%

\paragraph{Importance of retrieval query and subquestion generation.}
To demonstrate the importance of the retrieval query type and the graph-guidance, we conduct additional experiments with different retrieval queries and subgraph generation approaches in Table~\ref{tab:ablation}. 
The table shows that using the sub-questions as the retrieval query substantially outperforms using the input question. 
This demonstrates that the input question is insufficient to retrieve full knowledge to solve the multi-step reasoning problems.
For the subgraph generation, (w/o Subq Gen.) denotes the standard CoT without subgraph generation, and (w/o G-Guidance) decomposes the question into sub-questions only with an in-context learning scheme with a few demonstrations. 
The table shows that the graph-guidance consistently helps the model reason over multi-hop questions by guiding the reasoning process on different retrieval query types.
When using the sub-question as a retrieval query, the graph-guidance improves the prompting without graph-guidance with a margin of 3.67 F1 score on 2WikiMHQA dataset.
It leads to that the graph guidance is more effective in decomposing the question into sub-questions compared to the question decomposition, prompting only with a few exemplars.

\paragraph{Contribution of structural knowledge refinement and filtering.}
We provide the ablation studies to explore the contribution of structural knowledge refinement and filtering in Table~\ref{tab:abl2}.
The table shows that both structural knowledge refinement and filtering contribute to the performance improvement of our GE-Reasoning.
Especially, the structural knowledge refinement achieves the significant performance improvement of 1.4 on EM metric in 2WikiMHQA dataset.
This result indicates that the retrieved knowledge paragraphs contain irrelevant information in many cases and the refinement process can mitigate the problem.

\vspace{-4mm}
\paragraph{Performance comparison without knowledge retrieval.}
We also compare the performance of our prompting approach with other prompting methods without using knowledge retrieval (\textit{i.e., } closed book setting.)
Table~\ref{tab:noret} shows that our method achieves the best performance, indicating our propose method is effective in both the retrieval augmented setting and the pure LLM setting for multi-step reasoning.

\vspace{-4mm}

\paragraph{Quality of the constructed question graph.}
To empirically prove that our question graph construction generates an accurate question graph and denoting type improves the quality of the question graph and sub-questions based on the question graph, we evaluate the quality of the question graph and question-answering with and without type on 2WikiMHQA dataset using Llama-3 8B in Table~\ref{tab:graphqual}.
We evaluate the quality of the question graph with ground-truth graph of 2WikiMHQA. 
The table shows that our question graph construction accurately generates a question graph with and without type even using the smallest LLM we tried.
Also, it demonstrates using the type noticeably improves the reasoning capability of the prompting method with the 0.8 performance gain on EM score.

\subsection{Qualitative Analysis}
Here, we perform qualitative analysis by comparing the reasoning steps and answers generated by prompting without and with graph guidance in Table~\ref{tab:qual}.
(w/o G-Guidance) decomposes the question into sub-questions only with in-context learning without the graph guidance. 
Given the question ``Which film whose director is younger, Running for Grace or Irulum Oliyum?", (w/o G-Guidance) generates irrelevant sub-question ``Who directed Who is Running?" while the prompting with graph-guidance generates sub-questions and their answers relevant to solve the main question.
Rather, (w/o G-Guidance) gives the answer ``Who is Running", instead of ``Running for Grace" or ``Irulum Oliyum".
This case shows that the question decomposition without the guidance is prone to generating the wrong sub-question and our graph-guidance addresses it and effectively helps LLMs reason on multi-hop questions.


\section{Conclusion}
We have proposed a GE-Reasoning method to explicitly guide the large language models to reach the correct answer. 
We repeat the sub-question generation, answer generation, and answer filtering steps until predicting the final answer.
We use retrieval augmentation using intermediate sub-questions as queries to obtain the external knowledge triplets helpful for the intermediate reasoning processes.
Our experimental results on three multi-hop question answering benchmark datasets demonstrate the effectiveness of our GE-Reasoning methods.
\section{Limitations}
Similar to other prompting methods, the performance of our GE-Reasoning method relies on the large language models and the quality of the demonstrations.
In the open-book setting, the quality of retrieved knowledge is highly dependent on BM25 retriever. 
Therefore, using advanced retrieval methods help our model improve the performance.
\section{Ethics Statement}
Our GE-Reasoning prompting addresses the potential ethical issues of the large language models, such as the hallucination issue.
Some remaining concerns are that it could suffer from the ethical issue of the large language models such as Llama-3 and GPT-3.5 since it depends on them to reason on multi-hop questions.  

\bibliography{anthology,main}

\clearpage
\appendix

\section{Datasets and Licenses}
We use following three benchmark multi-hop question answering datasets to evaluate the performance of the baselines and our method.
To the best of our knowledge, these datasets do not have any privacy issue.
\begin{itemize}
    \item \textbf{2WikiMultihopQA}\footnote{Copyright (c) 2020 Xanh Ho, Licensed under Apache-2.0 license}~\cite{ho2020constructing} consists of 2-hop complex questions requiring the composition or comparison.
    \item \textbf{MuSiQue}\footnote{Copyright (c) Licensed under Apache-2.0 license}~\cite{trivedi2022musique} is a more challenging dataset where the problems include 2 to 4 hop questions that can be decomposed into simpler questions.
    \item \textbf{Bamboogle}\footnote{Copyright (c) 2022 Ofir Press, Licensed under MIT license}~\cite{trivedi2022musique} is a dataset consisting of 125 multi-hop questions where the supporting evidence is from Wikipedia.
\end{itemize}

\section{Implementaion Details}
Due to the heavy computational costs, we perform experiments with a single run on 2WikiMHQA and Musique datasets and 4 runs on Bamboogle dataset.
All the experiments are conducted using in-context learning~\cite{brown2020language} with 6 demonstrations to predict the answer on all the datasets.
For the knowledge retrieval, we employ BM25~\cite{robertson2009probabilistic} implemented with Elasticsearch and use top-2 retrieved documents following other RAG works~\cite{jiang2023active}.

\section{Additional Experiments}
We conduct additional experiments to compare the efficiency of our GE-Reasoning with other baseline prompting methods in Table~\ref{tab:efficiency}.
The time is measured as seconds per input question.
From the table, our GE-Reasoning shows the best performance on the 2WikiMHQA dataset with a comparable time compared to other retrieval-based methods.
In particular, GE-Reasoning improves 10.29 F1 score and 5.0 seconds per question compared to FLARE.
\begin{table}[!t]
    \small
    
    \centering
    \begin{tabular}{lccc}
        \toprule
        
        
          Methods    & EM & F1     & Time \\
         \midrule
         No retrieval& 31.6 & 38.94 & \textbf{2.8} \\
         One retrieval & 35.0 & 44.66  & 8.2 \\
         Verify-and-Edit &  36.4 & 43.00   & 14.6 \\
         FLARE & 41.8 & 50.70  & 21.9 \\
         IRCoT &  45.6 & 57.35  & 12.8 \\ 
         ERA-CoT &  35.2 & 42.93  & 11.2 \\ 
         \midrule
         \textbf{GE-Reasoning~(\textit{Ours})} & \textbf{53.0} & \textbf{60.99}  & 16.9 \\
        \bottomrule
    \end{tabular}
    \vspace{-2mm}
    \caption{Performance and efficiency comparisons on 2WikiMHQA dataset using Llama-3 8B. The time is evaluated as seconds per question. \textbf{Bold} indicates the best performance.
    }
    \label{tab:efficiency}
\end{table}

\section{Input Prompts}
\label{sec:appendix}
Table \ref{tab:pr2} shows the prompts for question graph construction.
For the subquestion and subanswer generation, we use prompts in Table~\ref{tab:pr1} and Table~\ref{tab:pr3}, respectively.
Table~\ref{tab:pr4} and Table~\ref{tab:pr5} are prompts for extracting triplets from the knowledge passages and filtering out irrelevant triplets in Section~\ref{sec:ret}.

    
    


\begin{table*}[!ht]
    
    

    \centering
    \small
    \begin{tabular}{p{14.4cm}}
        \toprule
Given a sentence, and all entities within the sentence. 
Extract all relationships between entities which directly stated in the sentence.
Every relationship stated as a triple: (E\_A; Relation; E\_B).\newline
Sentence: When did the director of film Hypocrite (Film) die?
Relation: (Hypocrite (Film); director; name: \#1), (name: \#1; death date; date: \#2)
\textcolor{white}{a}\newline
...\newline
\textcolor{white}{a}\newline
Given a sentence, and all entities within the sentence. 
Extract all relationships between entities which directly stated in the sentence.
Every relationship stated as a triple: (E\_A; Relation; E\_B).\newline
Sentence: \{sentence\}\newline
Triplets: \\
        \bottomrule
    \end{tabular}
    \caption{A question graph construction prompt for eliciting a graph from the question. 
    }
    \label{tab:pr2}
\end{table*}
\begin{table*}[!ht]
    
    

    \centering
    \small
    \begin{tabular}{p{14.4cm}}
        \toprule
Given the triplet, generate a subquestion based on the triplet.\newline
Triplet: (Hypocrite (Film); director; name: \#1)\newline
Subquestion: Who directed Hypocrite (Film)? \newline
\textcolor{white}{a}\newline
...\newline
\textcolor{white}{a}\newline
Given the triplet, generate a subquestion based on the triplet.\newline
Triplet: \{triplet\} \newline
Subquestion: \\
        \bottomrule
    \end{tabular}
    \caption{A subquestion generation prompt. 
    }
    \label{tab:pr1}
\end{table*}
\begin{table*}[!ht]
    
    

    \centering
    \small
    \begin{tabular}{p{14.4cm}}
        \toprule
Question: When did the director of film Hypocrite (Film) die?\newline
To answer this question, we answer the following subquestions:\newline
(1) Who directed Hypocrite (Film)?\newline
The film Hypocrite was directed by Miguel Morayta.\newline
(2) When did Miguel Morayta die?\newline
Miguel Morayta died on 19 June 2013.\newline
So the answer is 19 June 2013.\newline
\textcolor{white}{a}\newline
...\newline
\textcolor{white}{a}\newline
Question: \{question\}\newline
To answer this question, we answer the following subquestions:\newline
\{subquestion\} \\
        \bottomrule
    \end{tabular}
    \caption{A subanswer generation prompt. 
    }
    \label{tab:pr3}
\end{table*}
\begin{table*}[!ht]
    
    

    \centering
    \small
    \begin{tabular}{p{14.4cm}}
        \toprule
Extract triplets from the following paragraph:\newline
Maheen Khan is a Pakistani fashion and costume designer, also an award winner fashion designer for fashion labels \newline
\textcolor{white}{a}\newline
...\newline
\textcolor{white}{a}\newline
Triplets: \newline
(Maheen Khan; nationality; Pakistan)\newline
(Maheen Khan; profession; fashion and costume designer)\newline
(Maheen Khan; award winner; The Embroidery HouseMaheen)\newline
\textcolor{white}{a}\newline
...\newline
\textcolor{white}{a}\newline
Extract triplets from the following paragraph:\newline
\{paragraph\} \newline
Triplets: \\
        \bottomrule
    \end{tabular}
    \caption{A prompt for extracting triplets from the knowledge passage. 
    }
    \label{tab:pr4}
\end{table*}
\begin{table*}[!ht]
    
    

    \centering
    \small
    \begin{tabular}{p{14.4cm}}
        \toprule
Given knowledge triplets and a question, select triplets that are relevant to the question.\newline
Triplets:\newline
\{triplets extracted from the knowledge passage\}
Question:\newline
\{question\} \newline
Filtered triplets:\\
\bottomrule
    \end{tabular}
    \caption{A prompt for filtering out irrelevant triplets. 
    }
    \label{tab:pr5}
\end{table*}

\begin{algorithm*}
\caption{Algorithm of GE-Reasoning}
\label{alg}
\begin{algorithmic}[1]
\Require $x$: input question, $k_1, k_2$: hyper-parameter
\Ensure $\text{Answer}$: answer of the input question
\State $\mathcal{G}_{\boldsymbol{x}} \gets \text{ConstructGraph}\left(\boldsymbol{x}\right)$
\State $\text{Stop} \gets \text{False}$
\State $j \gets 1$
\State $\mathcal{C}_{\boldsymbol{x}} \gets \text{FindTripletswithOneVariable}\left(\mathcal{G}_{\boldsymbol{x}}\right)$
\While{$\text{Stop is False}$}
\State $\boldsymbol{t}^{(j)} \gets \text{Sample}\left(\mathcal{C}_{\boldsymbol{x}}\right)$
\State $\boldsymbol{q}^{(j)} \gets \text{SubquestionGenerate}\left(\boldsymbol{t}^{(j)}\right)$
\State $\boldsymbol{r}^{(j)} \gets \text{SubanswerGenerate}\left(\boldsymbol{q}^{(1)},\dots \boldsymbol{q}^{(j-1)},\boldsymbol{r}^{(j-1)}, \boldsymbol{q}^{(j)}\right)$
\If{$\text{Confidence}(\boldsymbol{r}^{(j)}) < k_1$}
    \State $\mathcal{D}_{{\qb}^{(j)}} \gets \text{Ret}\left({\qb}^{(j)} \right)$
    \State $\tilde{\mathcal{D}}_{{\qb}^{(j)}} \gets \text{KnowledgeRefine}\left(\mathcal{D}_{{\qb}^{(j)}}, {\qb}^{(j)} \right)$
    \State $\boldsymbol{r}^{(j)} \gets \text{SubanswerGenerate}\left(\tilde{\mathcal{D}}_{{\qb}^{(j)}}, \boldsymbol{q}^{(j)}\right)$
\EndIf
\If{$\text{Confidence}(\boldsymbol{r}^{(j)}) < k_2$}
    \State $\boldsymbol{q}^{(j)}, \boldsymbol{r}^{(j)} \gets ``", ``"$
\Else
    \State $\mathcal{G}_{\boldsymbol{x}} \gets \text{VariableAssignment}\left(\mathcal{G}_{\boldsymbol{x}},  \boldsymbol{r}^{(j)}, \boldsymbol{t}^{(j)}\right) $
\EndIf
\State $j \gets j + 1$
\State $\mathcal{C}_{\boldsymbol{x}} \gets \text{FindTripletswithOneVariable}\left(\mathcal{G}_{\boldsymbol{x}}\right)$
\If{$j >= 10 \text{ or } \mathcal{C}_{\boldsymbol{x}}=\emptyset$}
    \State$\text{Stop} \gets \text{True}$
\EndIf
\EndWhile
\State $\text{Answer} \gets \text{FinalAnswer}\left(\boldsymbol{q}^{(1)},\dots \boldsymbol{q}^{(j-1)},\boldsymbol{r}^{(j-1)}, \boldsymbol{q}^{(j)}, \boldsymbol{r}^{(j)} \right)$
\State \textbf{return} $\text{Answer}$
\end{algorithmic}
\end{algorithm*}

\end{document}